\documentclass[a4paper]{article}

\usepackage{graphicx}
\usepackage{subfigure}
\usepackage{booktabs} % for professional tables
\usepackage[a4paper, margin=3cm]{geometry}

\usepackage{hyperref}

\usepackage{multirow}
\usepackage{amsmath}
\usepackage{amssymb}
\usepackage{cellspace}
\usepackage{calc}
\usepackage{tikz}
\usepackage{pgfplots}
\pgfplotsset{compat=1.16}
\newcommand\cincludegraphics[2][]{\raisebox{-0.3\height}{\includegraphics[#1]{#2}}}
\DeclareMathOperator*{\argmin}{argmin}

\DeclareMathOperator{\logit}{logit}
\definecolor{magenta}{rgb}{0.8,0.0,0.8}
\definecolor{blue}{rgb}{0.4,0.4,0.8}
\definecolor{cyan}{rgb}{0.0,0.8,0.8}

\usepackage{algorithmic}

\usepackage{authblk}
\title{AVAE: Adversarial Variational Auto Encoder\footnote{pre-print version of an article to appear in the proceedings of the International Conference on Pattern Recognition (ICPR 2020) in January 2021}}
\author[1,2]{Plumerault, Antoine}
\author[1]{Le Borgne, Hervé}
\author[2]{Hudelot, Céline}
\affil[1]{Université Paris-Saclay, CEA, List, F-91120, Palaiseau, France \\ Email: \{antoine.plumerault,herve.le-borgne\}@cea.fr}
\affil[2]{MICS, Centrale-Supelec,  Gif-sur-Yvette, France \\Email: celine.hudelot@centralesupelec.fr}
\date{}                     %% if you don't need date to appear
\begin{document}


% find helpful for describing your paper; these are ugradientsed to populate
% \icmlkeywords{Generative Model, GAN, VAE}

%\author{\IEEEauthorblockN{Antoine Plumerault}
%\IEEEauthorblockA{Université Paris-Saclay, CEA, List,\\ F-91120, Palaiseau, %France\\MICS, Centrale-Supelec, \\ Gif-sur-Yvette, France\\
%Email: antoine.plumerault@cea.fr}
%\and
%\IEEEauthorblockN{Hervé Le Borgne}
%\IEEEauthorblockA{Université Paris-Saclay, CEA, List,\\ F-91120, Palaiseau, %France \\
%Email: herve.le-borgne@cea.fr}
%\and
%\IEEEauthorblockN{Céline Hudelot}
%\IEEEauthorblockA{MICS, Centrale-Supelec, \\ Gif-sur-Yvette, France\\
%Email: celine.hudelot@centralesupelec.fr}}

\maketitle
\begin{abstract}
Among the wide variety of image generative models, two models stand out: Variational Auto Encoders (VAE) and Generative Adversarial Networks (GAN). GANs can produce realistic images, but they suffer from mode collapse and do not provide simple ways to get the latent representation of an image. On the other hand, VAEs do not have these problems, but they often generate images less realistic than GANs. In this article, we explain that this lack of realism is partially due to a common underestimation of the natural image manifold dimensionality. To solve this issue we introduce a new framework that combines VAE and GAN in a novel and complementary way to produce an auto-encoding model that keeps VAEs properties while generating images of GAN-quality. We evaluate our approach both qualitatively and quantitatively on five image datasets.
\end{abstract}
\section{Introduction}
Since the original GAN paper \cite{GAN}, generative models have successfully leveraged the power of deep learning to generate complex data distribution with increasing fidelity. Generative models are now used for a wide variety of tasks, including notably sample generation but also photo manipulation \cite{IAN}, style transfer \cite{CYCLEGAN}, pre-processing for face recognition \cite{TPGAN}, text to image translation \cite{STACKGAN} and controlled image generation~\cite{STEERABILITY3}.   

In the literature, two families of generative models stand out for image data: Variational Auto Encoders (VAE) \cite{VAE} and Generative Adversarial Networks (GAN) \cite{GAN}, each exhibiting respective advantages and limitations. GANs usually produce more realistic images \cite{BIGGAN, STYLEGAN} but they are notoriously difficult to train and suffer from mode collapse \cite{UNROLLEDGAN}. Moreover, when using GANs, there is no trivial way to get the latent representation of an image, limiting their use. In contrast, VAE models do not share these problems but the images they generate suffer from a lack of realism. It is often explained by the use of inappropriate reconstruction errors. Some previous works \cite{VAEGAN, BIGAN, ALI} have proposed solutions to solve these problems by combining or modifying these two frameworks. However, these methods exhibit a trade-off between the realism of the generated images and the fidelity of the reconstructions. In this paper, we show that GANs and VAEs can be complementary in the sense that we can derive two complementary losses from them. From this observation, we propose the AVAE model which is a VAE style model to produce samples of comparable quality as those generated by a GAN while allowing high fidelity reconstructions when used as an auto-encoder. In comparison to \cite{VAEGAN} who first introduces the idea of a combination of the two frameworks, we provide theoretical insights to show the pertinence of our approach and we address the problem of the trade-off between realism and reconstruction accuracy.

The paper is organized as follows. We begin with a reminder of GAN and VAE frameworks and explain their limitations. Then we investigate how they can be combined effectively. We thus propose an effective approach to do so, named AVAE. At last, we present a qualitative and quantitative evaluation of the performance of our model on a variety of image datasets comparing it with the state of the art. We also show that our method scales well to high resolution images.
\section{Background}
\subsection{Variational Auto Encoders}\label{sec:VAE}
VAE \cite{VAE} is a framework to learn deep latent variable models. It assumes that observed data $X$ result from random variables $z\sim p(z)$ in a latent space $\mathcal{Z}$ such that it exists a deterministic function $f: (z, \epsilon) \rightarrow x$, $\epsilon$ being a stochastic noise. The probability of observing $x$ knowing $z$ is estimated by a \emph{decoder} model $p_{\theta_d} : z \mapsto p_{\theta_d}(x|z)$ parametrized by $\theta_d$ and on the contrary, the probability that $z$ is the latent source of $x$ is estimated by a \emph{encoder} model $q_{\theta_e} : x \mapsto q_{\theta_e}(z|x)$ parametrized by $\theta_e$. To estimate the parameters of the generative model of the data $X = (x^{(1)}, ..., x^{(N)})$ with $N$ the number of observed samples, we maximize the log likelihood of the observations: $\log{p_{\theta_d}\left(x^{(i)}\right)} = \log{\int_\mathcal{Z} p_{\theta_d}\left(\left.x^{(i)}\right|z\right)p(z)dz}$. Computing $\log{p_{\theta_d}\left(x^{(i)}\right)}$ is nevertheless intractable in practice, thus \cite{VAE} proposes to maximize a tractable lower bound, leading to the following loss to train the VAE:
\begin{equation}
\begin{aligned}
    \mathcal{L}_\text{VAE}\left(\theta_e, \theta_d; x\right) 
    ={}& \underbrace{\mathbb{E}_{q_{\theta_e}\left( z \vert x \right)}\left[ -\log{ p_{\theta_d}\left(x \vert z \right)}\right] }_{\mathcal{L}_\mathcal{R}}\\
    &+ \underbrace{\text{KL} \left(\left. q_{\theta_e}\left( z \vert x \right) \right\Vert p(z) \right)}_{\mathcal{L}_\mathcal{P}}
\end{aligned}
\end{equation}
with $p_{\theta_d}$ usually chosen as a Gaussian distribution $\mathcal{N}(x;\mu_{\theta_d}(z),Id)$ and $\text{KL}$ the Kullback-Leibler divergence. Hence, the term $\mathcal{L}_\mathcal{R}=\mathbb{E}_{q_{\theta_e}\left( z \vert x \right)}\left[-\log{ p_{\theta_d}\left(x \vert z \right)}\right]=\mathbb{E}_{q_{\theta_e}\left( z \vert x \right)} \left[ \frac{1}{2}\left\Vert \mu_{\theta_d}(z)-x\right\Vert^2  \right]$ can be interpreted as a \textit{reconstruction} error and is estimated by Monte-Carlo method (usually with a single sample), and the term $\mathcal{L}_\mathcal{P}=\text{KL} \left(\left. q_{\theta_e}\left( z \vert x \right) \right\Vert p(z) \right)$ forces the distribution of the latent space to match the \textit{prior} $p(z)$. Usually, $p(z)$ is a standard Gaussian distribution $\mathcal{N}(z; 0,Id))$. The $\mathcal{L}_\mathcal{P}$ term acts as an information bottleneck on the latent produced by the encoder. Indeed:
\begin{equation}
\begin{aligned}
    \mathbb{E}\left[\mathcal{L}_\mathcal{P}\right] &= \mathbb{E}\left[\text{KL}\left( q_{\theta_e}\left( z \vert x \right) \Vert p(z) \right) \right]\\
    &= \sum_{i=1}^N p(x^{(i)}) \int_{\mathcal{Z}} q_{\theta_e}(z|x^{(i)}) \log{\frac{q_{\theta_e}(z|x^{(i)})}{p(z)}} dz \\
    &= \sum_{i=1}^N \int_{\mathcal{Z}} p_{\theta_e}(z,x^{(i)}) \log{\frac{p_{\theta_e}(z,x^{(i)})}{p(x^{(i)})p(z)}} dz \\
    &= \sum_{i=1}^N \int_{\mathcal{Z}} p_{\theta_e}(z,x^{(i)}) \log{\frac{p_{\theta_e}(z,x^{(i)})p_{\theta_e}(z)}{p(x^{(i)})p(z)p_{\theta_e}(z)}} dz\\
    &= \sum_{i=1}^N \int_{\mathcal{Z}} p_{\theta_e}(z,x^{(i)}) \log{\frac{p_{\theta_e}(z,x^{(i)})}{p(x^{(i)})p_{\theta_e}(z)}} dz \\
    &~+ \int_{\mathcal{Z}} p_{\theta_e}(z) \log{\frac{p_{\theta_e}(z)}{p(z)}} dz\\
    &= I_\theta(x; z) + \text{KL}(p_{\theta_e}(z)||p(z))\\
\end{aligned}
\end{equation}
with I the mutual information between $x$ and $z$. This term thus limits the amount of information about the original image that goes through the latent code and pushes the distribution of the latent code produced by the encoder to match the prior latent code distribution.
\subsubsection*{Limitations of the VAE framework}
\begin{figure}[t]
    % \vskip 0.1in
    \begin{center}
    \includegraphics[width=0.9\columnwidth]{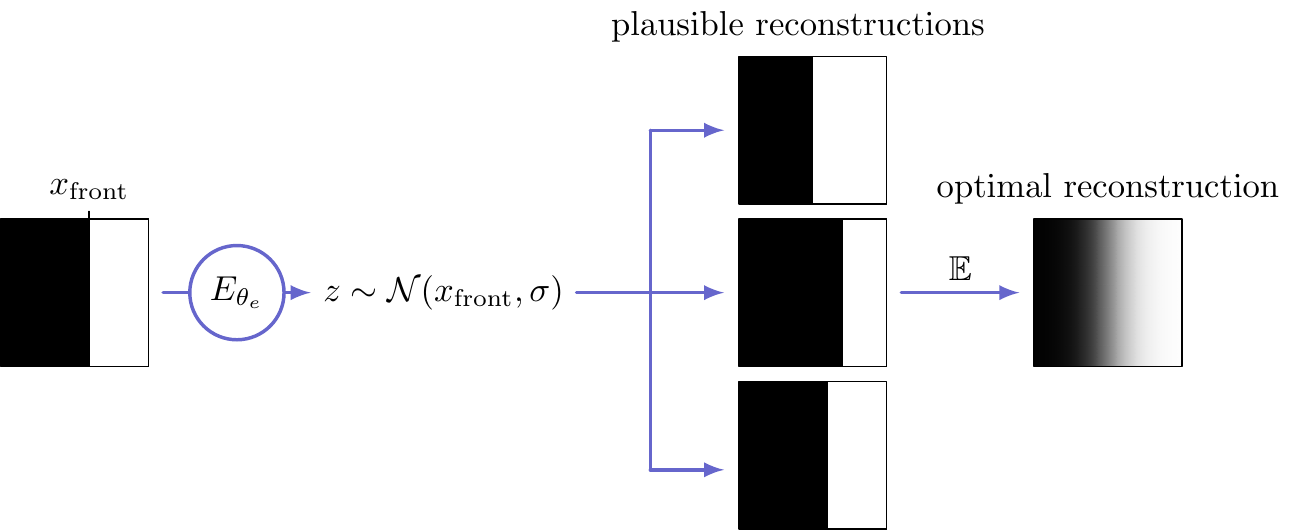}
    \vskip -0.1in
    \caption{Illustration on why VAEs produce blurry reconstructions. Consider the example of a binary frontier in an image $i$ and a latent code $z$ which corresponds to the position of the frontier $x_\text{front}$. If $q_{\theta_e}(z|i) = \mathcal{N}(x_\text{front}, \sigma)$ then $p_{\theta_e}(x_\text{front} \vert z) = \mathcal{N}(z, \sigma)$ and the optimal reconstruction of the pixel at position $x$ is $\mathbb{E}\left[pixel(x) \vert z \right] = 1 \times \mathbb{P}_{\theta_e}(x>z) + 0 \times \mathbb{P}_{\theta_e}(x<z) = \frac{1}{2}\left(1+\text{erf}(\frac{x-z}{\sqrt{2}\sigma})\right)$ which is a smooth transition between black and white instead of a sharp transition in the original binary image.}
    \label{fig:front}
    \end{center}
    \vskip -0.3in
\end{figure}
Understanding and improving VAE are active subjects of research. Some works have focused on reducing the gap in quality which often exists between reconstructions produced by VAEs and images sampled with them \cite{Diagnosing_and_Enhancing_VAE_Models}. Others have aimed at learning more interpretable latent space structure \cite{Learning_Hierarchical_Priors_in_VAEs}. While dealing with interesting issues, these papers are not in line with the problem tackled in this paper related to the lack or realism and the blurry aspect of images generated or reconstructed with VAEs.
% However, the aspect we want to tackle in this paper is that, when training a VAE on images, reconstructions and generated images produced by VAE models are often blurry and not visually realistic. This blurriness can be explained by several factors.

As we have seen, $\mathcal{L}_\mathcal{P}$ acts as an information bottleneck which limits the information about the original image $x$ that passes through the latent code. This creates an uncertainty on the attributes of the original image $x$ when trying to reconstruct it. This uncertainty combined with the use of the mean square error as a reconstruction error causes the generated images $\mu_{\theta_d}(z)$ to be blurry. Indeed under those circumstances, the optimal value for each pixel of the reconstructed image is its expected value given the information available in the latent code \cite{BLURRYL2} (See Figure~\ref{fig:front} for an illustration). 

The second aspect that prevents VAE and AE in general to produce realistic samples is the use of a pixel-wise reconstruction error combined with the high dimensionality of the natural image manifold. Indeed, it is often assumed that natural images lie on a low dimensional manifold of image space, in particular because of a strong redundancy at a local scale~\cite{kretzmer52tv_signals}. This point is globally asserted by empirical evidence~\cite{ruderman94stat_img_nat} but can be mitigated with regard to textures. We argue that textures like wood, hair or waves of the ocean are living in a much higher dimensional manifold. This manifold thus cannot be captured in the low dimensional latent space of generative models even in the absence of an explicit information bottleneck. Indeed, it would require a network with a very high capacity to map the low dimensional latent into a high dimensional manifold. One can convince himself of this fact by considering that hair configuration is the product of the configuration of each individual strand of hair. GANs can partially overcome this problem with mode collapse on textures by generating only a subset of this manifold which is enough to fool the discriminator network. However, the use of a powerful pixel-wise reconstruction error in the case of VAE prevents the decoder from using this strategy leading to unrealistic results.
\subsection{Generative Adversarial Networks}
GAN globally consists in training two neural networks with adversarial objectives to generate samples indistinguishable from the samples taken from the dataset. The \textit{generator} network parametrized by $\theta_g$ is trained to map a random vector to the data space. The \textit{discriminator} or \textit{critic} network parametrized by $\theta_c$ is a classifier that is trained to distinguish real samples from generated ones. The key point is that the generator does not have access to real data and can only improve its parameters through its ability to fool the discriminator. The objective of the critic is:
\begin{equation}
\begin{aligned}
    \mathcal{O}_C(\theta_c) ={}& \mathbb{E}_{x \sim p(x)}\left[
        \log{\left(1-C_{\theta_c}(x)\right)}
    \right] \\
    & + \mathbb{E}_{x \sim p_{\theta_g}\left(x \vert z \right)}\left[
        \log{C_{\theta_c}(x)}
    \right]
\end{aligned}
\end{equation}
while the generator tries to fool the critic by minimizing:
\begin{equation}
    \mathcal{O}_G(\theta_g) = 
    \mathbb{E}_{x \sim p_{\theta_g}\left(x \vert z \right)}\left[
        \log{\left(1 - C_{\theta_c}(x)\right)}
    \right]
\end{equation}
\subsubsection*{Limitations of the GAN framework}
GANs have proven to be very successful for generation tasks but suffer from two major limitations in comparison to VAEs: mode collapse and the absence of an encoder network. \emph{Mode collapse} occurs when, at each step, the generator is able to only produce a few different samples. In its extreme case, the generator only produces one type of sample, that is thus easily recognized by the discriminator. In return, the discriminator does not need real data to train and its feedback to the generator through back-propagation does no longer contain useful information. More commonly, the generator produces a limited number of samples and interpolation of them. 

Even when a GAN appears to have attained a good solution, \emph{mode collapse} may have occurred slightly and some modes of the data distribution may be missed by the generator. \emph{Mode collapse} also raises the question of the existence of an acceptable pseudo-inverse mapping of the generator defined on the entire dataset space. 
The second issue is that the GAN framework does not provide an explicit model to find the latent space fibers of samples as it does not have an encoder. 
\section{Related Works}
To leverage both the advantages of GANs and VAEs, \cite{VAEGAN} proposed the VAE/GAN architecture which combines them. They propose to add a discriminator to push reconstructions from the VAE toward more realism and replaced the standard reconstruction error by a perceptual similarity metric based on the filters learned by the discriminator. This approach is problematic because the discriminator is trained to predict whether an image is a real one or a fake one. Thus, the features extracted from it may not be adapted to describe image content making them a disputable choice to base a similarity metric on. As an example, we noticed that VAE/GAN sometimes fails to reconstruct precisely skin color on the CelebA dataset (see Figure~\ref{fig:qualitative_rec}) as this information might be useless to some extent for the discriminator. If carefully tuned, this approach tends to work well in practice and allows sharper reconstructions. Nevertheless, \cite{ALI} pointed out that this approach also tends to exhibit a compromise between VAE and GAN and produces less realistic samples than GAN. They propose the BiGAN architecture~\cite{BIGAN} which is composed of an encoder that transforms real images into latent codes, a generator that transforms latent codes sampled randomly into images and a discriminator which tries to guess the origin of a couple of image/latent. While this approach is very elegant and produces samples of the same quality as GANs, it is aimed at finding good feature representations in an unsupervised way and often fails to produce very accurate reconstructions. In \cite{SVAE} and \cite{ALICE}, the authors propose variations of the BiGAN framework and additional theoretical insights about the latter. They produce more accurate reconstructions in terms of MSE but they are blurry (no hair texture when trained on faces images) which is precisely the issue we aim at solving here. In \cite{IAN}, the authors propose a variation of the VAE/GAN framework where the encoder and the discriminator network are a unique model. While it is not clear why this choice is a good one or not, the model reconstruction loss is the combination between a pixel-wise error and the VAE/GAN reconstruction loss which introduces a compromise between the blurriness of the reconstructions and the features reconstruction fidelity. Similarly, \cite{IntroVAE} have proposed an elegant framework where the discrimination is made on the latent space. Our approach introduces a reconstruction loss that does not interfere with the realism of the images while being linked with the MSE. By combining our reconstruction loss with adversarial training, we are able to produce photo-realistic reconstructions with no compromise on fidelity. Moreover, our framework is theoretically grounded and is not limited to image data as we show on a toy example (Section~\ref{sec:toy}) that it can be used in a more general context.

A recent work published at ECCV 2020 proposed an architecture very similar to ours~\cite{highfidelitywithdisantangledrepresentation}, but differs from our work on several crucial points. First, the reconstruction loss at the core of our contribution is different, due to the theoretical aspects developed in the following. We were primarily interested in the theoretical aspects behind the realism of generated images while their work focuses more on the disentangled properties of the framework. Our evaluation method also differs on the choice of datasets and metrics. Overall, we believe that our works complement each other well. 

\section{The AVAE framework}
\subsection{Complementarity between VAE and GAN}\label{sec:complementarity}
\begin{figure}[t]
    \begin{center}
    \def\arraystretch{0.0}
    \setlength\tabcolsep{0pt}
    \begin{tabular}{cccc}
    \multicolumn{4}{c}{\includegraphics[width=0.9\columnwidth]{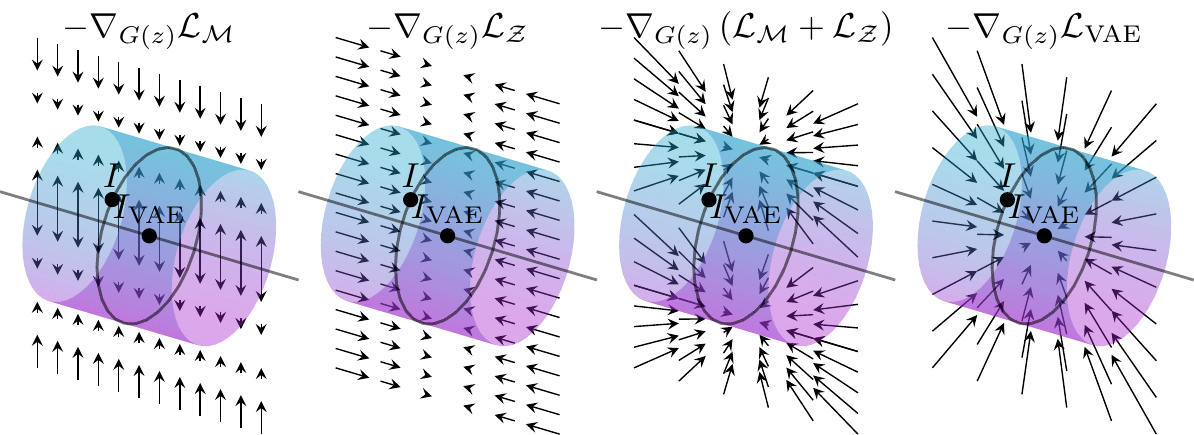}} \\
    \parbox{0.225\columnwidth}{\centerline{(a)}} & 
    \parbox{0.225\columnwidth}{\centerline{(b)}} & 
    \parbox{0.225\columnwidth}{\centerline{(c)}} & 
    \parbox{0.225\columnwidth}{\centerline{(d)}}
    \end{tabular}
    \caption{The figure depicts a small portion of data space. The cylinders symbolize the real data high-dimensional manifold, the black line the low-dimensional manifold on which the reconstructions of VAEs are restricted. The images on the black circle where the image $I$ is located are all mapped to the same latent code by the encoder network. Thus, they share a common reconstruction: $I_\text{VAE}$. This reconstruction is outside of the data manifold as it is the expected value of the original image given the latent code computed by the encoder which is blurry. The arrows represent the gradient of different losses (w.r.t the reconstruction) that are minimized during training: (a) the loss derived from the GAN framework that pushes the reconstructions toward the data manifold, (b) the loss derived from the VAE framework that pushes the reconstructions toward a region where images are mapped to the same latent code by the encoder, (c) their combination and (d) the VAE reconstruction loss i.e. the mean square error. (Note that gradients are represented on a single plane, while there is a radial symmetry around the black line)}
    \label{fig:losses}
    \end{center}
    \vskip -0.3in
\end{figure}

Despite their differences, we show that VAE and GAN exhibit some form of complementarity and that we can build a hybrid approach that solves several problems listed above. One naive hybridization could be to train a VAE with an additional adversarial loss term to push reconstructions toward more realism. However, as we have seen, optimal reconstructions are not always realistic. This approach would lead to choosing a trade-off between reconstruction accuracy and realism as both have conflicting objectives. One of the contributions of this paper is to show that we can derive two complementary losses from the VAE and GAN frameworks which share an optimal solution allowing accurate and realistic reconstructions.
In the GAN framework, we can derive a \textit{manifold} loss $\mathcal{L}_\mathcal{M}$ from the discriminator network which judges the realism of a given sample. This loss can be interpreted as a ``distance'' between the data manifold and a sample as described in \cite{EBGAN}. 
In the VAE framework, we train an encoder which maps data in a latent space $\mathcal{Z}$. This latent space can be seen as a map of the data manifold. Distances in the latent space can be interpreted as a distance between two points of the data manifold. This loss is noted $\mathcal{L}_\mathcal{Z}$. Our intuition, depicted by Figure~\ref{fig:losses}, is that these two losses can be used in conjunction to train a model which produces realistic images while keeping approximately the latent space organization of a VAE.

We give here further explanation on why the VAE framework fails to produce realistic images and what conditions a reconstruction error should satisfy to achieve accurate and realistic reconstructions. Let us consider an auto encoder that uses a reconstruction error of the form $\mathcal{L}\left(x, y\right) = \left\Vert x - y\right\Vert^2$. Let us note $x$ the input, $z$ the output of the encoder $E_{\theta_e}$ and $\hat{x}$ the output of the decoder $D_{\theta_d}$. With the parameters of the encoder fixed, the optimal reconstruction should minimize the expected cost over the potential images $\tilde{x}$ that could have produced the observed $z$. i.e.
\begin{equation}
\begin{aligned}
    \hat{x}^*(z) \in \argmin_{\hat{x}} \mathbb{E}_{\tilde{x} \sim p_{\theta_e} ( \tilde{x} \vert z )} \left[ 
        \left\Vert 
        \tilde{x} - \hat{x}
        \right\Vert^2
    \right] 
\end{aligned}
\end{equation}
Thus the optimal solution is given by $\hat{x}^*(z) = \mathbb{E}_{\tilde{x} \sim p_{\theta_e} ( \tilde{x} \vert z )} \left[ \tilde{x} \right]$.
%
% \begin{equation}
% \begin{aligned}
%     \hat{x}^*(z) = \mathbb{E}_{\tilde{x} \sim p_{\theta_e} ( \tilde{x} \vert z )} \left[ \tilde{x} \right]
% \end{aligned}
% \end{equation}
%
The problem is that, in this case the optimal reconstruction $\hat{x}^*$ is the expected value of all the possible reconstructions given the knowledge of the latent code. It leads to a blurry reconstruction, quite unlikely under the data distribution $p_\mathcal{D}$ (i.e. $p_\mathcal{D}(\hat{x}^*)$ is small).

In a more general setting we can consider objectives of the form: $L(\hat{x}, x) = ||f(\hat{x}) - g(x)||^2$ where $f$ is an arbitrary differentiable function and $g$ is a more general stochastic function. In this case, the optimal solution verifies:
\begin{equation}
\begin{aligned}
    f(\hat{x}^*(z)) = \mathbb{E}_{g(x) \sim p_{\theta_e} (g(x) \vert z )} \left[ g(x) \right]
\end{aligned}
\end{equation}
This objective has a common optimum with the GAN objective, if and only if we have $p(f(x^*(z))) = p(f(x))$ for $z \sim p(z)$ and $x \sim p_\mathcal{D}(x)$. However, to be what we can call a good reconstruction error, $f$ and $g$ should also carry the maximum of information about their input and be close to each other to make the loss able to discriminate between accurate and not accurate reconstructions.
\subsection{Architecture}
Similarly to VAE, the proposed AVAE framework is based on an encoder $E_{\theta_e}$ and a decoder $D_{\theta_d}$. We add two additional models: a generator $G_{\theta_g}$ and a critic $C_{\theta_c}$. The role of the generator is to produce realistic samples from latent codes. 

The VAE part of our framework is similar to classical VAE: it is a parametrized model $q_{\theta_e}(z|x) = \mathcal{N}\left(z; \mu_{\theta_e}(x), \Sigma_{\theta_e}\right)$ with $\Sigma_{\theta_e}$ a diagonal matrix of the form $\text{diag}(\sigma^2_{\theta_e})$. The prior distribution of the latent codes is $p(z) = \mathcal{N}\left(z; 0, Id\right)$ and $p_{\theta_d}(x|z) = \mathcal{N}\left(x; \mu_{\theta_d}(z), Id \right)$. With such choices, $\mathcal{O}_\text{VAE}(\theta_e, \theta_d; x)$ can be estimated by a Monte-Carlo method. Indeed, the Kullback-Leibler divergence term of the loss $\text{KL} \left( q_{\theta_e}\left( z \vert x \right) \Vert p(z) \right)$ is equal to:
% The VAE part of our framework is a parametrized model $q_{\theta_e}(z|x) = \mathcal{N}\left(z; \mu_{\theta_e}(x), \Sigma_{\theta_e}\right)$ with $\Sigma_{\theta_e}$ a diagonal matrix of the form $\text{diag}(\sigma^2_{\theta_e})$. The prior distribution of the latent codes is $p(z) = \mathcal{N}\left(z; 0, Id\right)$ and $p_{\theta_d}(x|z) = \mathcal{N}\left(x; \mu_{\theta_d}(z), Id \right)$. With such choices, $\mathcal{O}_\text{VAE}(\theta_e, \theta_d; x)$ can be estimated by a Monte-Carlo method. Indeed, the Kullback-Leibler divergence term of the loss $\text{KL} \left( q_{\theta_e}\left( z \vert x \right) \Vert p(z) \right)$ is equal to:
%
\begin{equation}
    \frac{1}{2} \sum_{j=1}^{\text{dim}(\mathcal{Z})}
    \sigma_{\theta_e j}^2 
    + \mu_{\theta_e}^2(x)_j - 1
    - \log{\sigma_{\theta_e j}^2} 
\end{equation}
The reconstruction term of the loss $\mathbb{E}_{q_{\theta_e}(z\vert x)}\left[\log{p_{\theta_d}(x \vert z)}\right]$ can be estimated by Monte-Carlo, sampling $z$ from $q_{\theta_e}(z\vert x)$ and noting that:
\begin{equation}
    \log{p_{\theta_d}(x|z)} = 
        -\frac{\text{dim}(x)}{2}\log{2\pi} - \frac{1}{2} \left\Vert\mu_{\theta_d}(z) - x\right\Vert^2 
\end{equation}
$z$  being sampled from $q_{\theta_e}\left(z \vert x \right)$, the loss of the VAE for one sample is the following (without constant terms):
\begin{equation}
\begin{aligned}
    \mathcal{L}_\text{VAE}(\theta_e&, \theta_d; x) 
        = \frac{1}{2} \left\Vert\mu_{\theta_d}(z) - x\right\Vert^2 \\
        &+ \frac{1}{2}\sum_{j=1}^{\text{dim}(\mathcal{Z})}
    \sigma_{\theta_e j}^2 
    + \mu_{\theta_e}^2(x)_j
    - \log{\sigma_{\theta_e j}^2}
\end{aligned}
\end{equation}
For the generator part, when we want to use it for reconstruction, we build its input by concatenating $z$ the latent code produced by the encoder with a random vector $\xi$ sampled from $\mathcal{N}(0, Id)$ to form the latent code for our generator. $z$ encodes the information captured by the encoder while $\xi$ encode the variation not captured by it. With this choices, we sample from $p_{\theta_g}(x \vert z)$ by taking $x = G_{\theta_g}(z, \xi)$. Note that $\xi$ can be removed if we consider that for a given $z$ there is only one possible reconstruction but we present here the general setting as we consider. To sample a random image from the generator we simply sample $z$ from the prior distribution defined in the VAE part and $\xi$ from $\mathcal{N}(0, Id)$. Ideally, the generator should invert the encoder and thus $p_{\theta_g}(x \vert z)$ should be as close as possible than $p_{\theta_e}(x \vert z)$. This consideration leads us to minimizing the following negative log likelihood with $z\sim \mathcal{N}(0, Id)$ and $x \sim p_{\theta_g}(x \vert z)$ :
% For the generator part, we concatenate $z$ with a random vector $\xi$ sampled from $\mathcal{N}(0, Id)$ to form the latent code for our generator, modeling sampling from $p_{\theta_g}(x \vert z)$ by taking $x = G_{\theta_g}(z, \xi)$. With this formulation, we expect $z$ to encode the information captured by the decoder and $\xi$ to encode complementary information. Note that $\xi$ can be removed but we present here the general setting. 
%
\begin{equation}
\begin{aligned}
    \mathcal{L}_G(\theta_g) ={}& \mathbb{E}\left[ -\log{p_{\theta_e}(x \vert z)}\right]\\
    ={}& \mathbb{E}\left[-\log{p_{\theta_e}(z \vert x) p(x)}\right] + C\\
    ={}& \mathbb{E}\left[-\log{p_{\theta_e}(z \vert x)}\right] + \mathbb{E}\left[\log{\frac{p_{\theta_g}(x)}{p(x)p_{\theta_g}(x)}}\right] + C \\
    ={}& \mathbb{E}\left[-\log{p_{\theta_e}(z \vert x)}\right] + \text{KL}(p_{\theta_g}(x) \Vert p(x)) + H_{\theta_g} + C
\end{aligned}
\end{equation}
with $H_{\theta_g}$ the differential entropy of the distribution $p_{\theta_g}(x)$. The term $\log{p_{\theta_e}\left(z \vert x \right)}$ can be computed directly:
\begin{equation}\label{eq:logp}
\begin{aligned}
    \log{p_{\theta_e}\left(z \vert x \right)}
    ={}& \log{\mathcal{N}(z; \mu_{\theta_e}(x), \Sigma_{\theta_e})} \\
    ={}& -\frac{\text{dim}(\mathcal{Z})}{2}\log{2\pi}- 
    \frac{1}{2} \log{\vert \Sigma_{\theta_e} \vert} \\ 
    &-\frac{1}{2} \left\Vert\frac{\mu_{\theta_e}(x) - z }{\sigma_{\theta_e}}\right\Vert^2 
\end{aligned}
\end{equation}
We define the reconstruction loss $\mathcal{L}_\mathcal{Z}$ by removing constant terms in Equation~\ref{eq:logp}:
\begin{figure}[t]
\begin{center}
\centerline{\includegraphics[width=0.8\columnwidth]{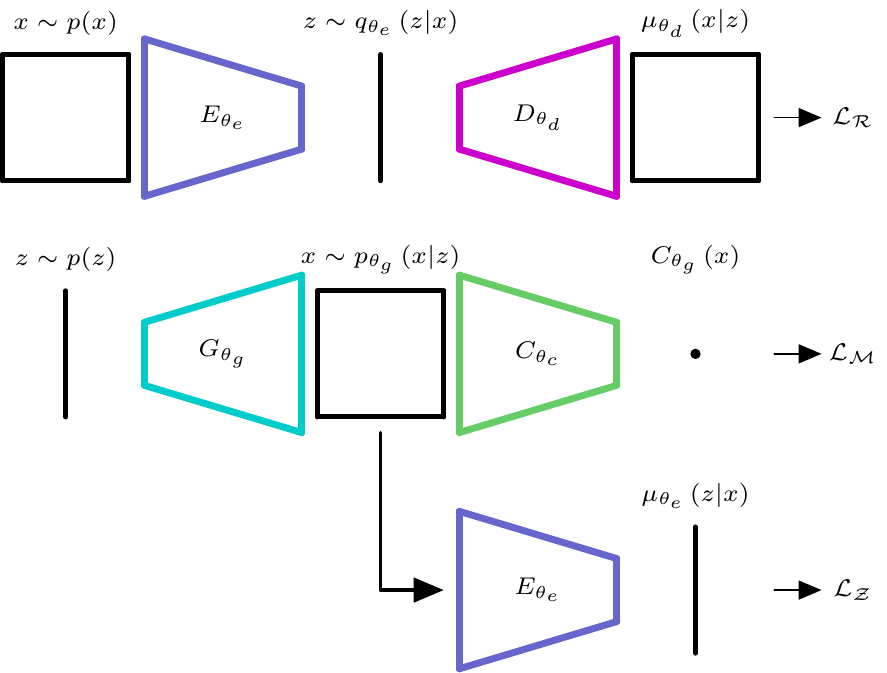}}
\vskip -0.1in
\caption{Summary of our Adversarial Variational Auto Encoder framework. $E_{\theta_e}, D_{\theta_d}, G_{\theta_g}, C_{\theta_c}$ are respectively the encoder, decoder, generator and critic (discriminator). Note that the weights of the encoder $E_{\theta_e}$ are shared between the two architectures. Images are denoted by the letter $x$ and latent codes by the letter $z$.}
\label{fig:framework}
\end{center}
\vskip -0.3in
\end{figure}
\begin{figure}[t]
\label{alg:method}
\begin{algorithmic}
    \STATE Initialize parameters of the models: $\theta_e$, $\theta_d$, $\theta_g$, $\theta_c$  \;
    \WHILE{training}
        \STATE
        \COMMENT{Forward pass.}
        \STATE $x^\text{real} \leftarrow$ batch of images sampled from the dataset.
        \STATE $z^\text{real}_\mu, z^\text{real}_\sigma \leftarrow E_{\theta_e}(x^{real})$
        \STATE $z^\text{real} \leftarrow z^\text{real}_\mu + \epsilon z^\text{real}_\sigma$ with $\epsilon \sim \mathcal{N}(0,Id)$
        \STATE $\mu^\text{real} \leftarrow D_{\theta_d}(z^\text{real})$
        \STATE $x^\text{fake} \leftarrow G_{\theta_g}(z^\text{fake}, \xi)$ with $z^\text{fake}, \xi \sim \mathcal{N}(0,Id)$
        \STATE $z^\text{fake}_{\mu}, z^\text{fake}_\sigma \leftarrow E_{\theta_e}(x^\text{fake})$
        \STATE $C^\text{real}, C^\text{fake} \leftarrow C_{\theta_c}(x^\text{real}), C_{\theta_c}(x^\text{fake})$
        \STATE
        \COMMENT{Compute losses gradients and update parameters.}
        \STATE $\theta_e \xleftarrow{-} \nabla_{\theta_e} \mathcal{L}_\text{VAE}(\theta_e, \theta_d)$ ; $\theta_g \xleftarrow{-} \nabla_{\theta_g} \mathcal{L}_G(\theta_g)$
        \STATE $\theta_d \xleftarrow{-} \nabla_{\theta_d} \mathcal{L}_\text{VAE}(\theta_e, \theta_d)$ ; $\theta_c \xleftarrow{-} \nabla_{\theta_c} \mathcal{L}_C(\theta_c)$
        % \STATE $\theta_g \xleftarrow{-} \nabla_{\theta_g} \mathcal{L}_G(\theta_g)$ ; 
        % \STATE $\theta_c \xleftarrow{-} \nabla_{\theta_c} \mathcal{L}_C(\theta_c)$
    \ENDWHILE
\end{algorithmic}
\vskip -0.1in
\caption{Algorithm to train the Adversarial Variational Auto Encoder.}
\vskip -0.2in
\end{figure}
\begin{equation}
    \mathcal{L}^a_\mathcal{Z}(\theta_g; z, \theta_e) =  
    \frac{1}{2} \left\Vert\frac{\mu_{\theta_e}(x) - z }{\sigma_{\theta_e}}\right\Vert^2
\end{equation}
We can estimate the second term by training a classifier $C$ that discriminates generated images from real ones by minimizing the cross-entropy:
\begin{equation}
\begin{aligned}
    \mathcal{L}_C(\theta_c) ={}& -\mathbb{E}_{x \sim p(x)}\left[
        \log{\left(1-C_{\theta_c}(x)\right)}
    \right] \\
    &- \mathbb{E}_{x \sim p_{\theta_g}\left(x \vert z \right)}\left[
        \log{C_{\theta_c}(x)}
    \right]
\end{aligned}
\end{equation}
Under this loss, the optimal solution for $C$ is: 
\begin{equation}
    C^*: x \rightarrow \frac{p_{\theta_g}(x)}{p(x)+p_{\theta_g}(x)}
\end{equation} 
$\mathcal{L}_{\mathcal{M}}$ is then defined by sampling $x$ from $p_{\theta_g}(x)$. Hence:
\begin{equation}
    \mathcal{L}_{\mathcal{M}}(\theta_g; x, \theta_e) = \logit{C} 
\end{equation}
Indeed, $\logit{C} \approx \logit{C^*} = \log\left(\frac{p_{\theta_g}\left(x\right)}{p(x)}\right)$ which is an unbiased estimator of the Kullback-Leibler divergence term. Minimizing the differential entropy $H_{\theta_g}$ of the distribution $p_{\theta_g}(x)$ will push it to be as peaked as possible and is not data dependent. Moreover, this term is intractable. Hence, as a form of regularization, we remove it. One problem still remains. Indeed the optimal reconstruction for $\mathcal{L}^a_\mathcal{Z}$ verifies the following equation: $\mu_{\theta_e}(\hat{x}^*(z)) = z$ and thus $p(\mu_{\theta_e}(\hat{x}^*(z))) = \mathcal{N}(\mu_{\theta_e}(\hat{x}^*(z));0, I)$ while $ p(\mu_{\theta_e}(x)) = \mathcal{N}(\mu_{\theta_e}(x);0, I-\Sigma)$. To solve this problem, we propose to replace the expression of $\mathcal{L}^a_\mathcal{Z}$ by:
\begin{equation}
    \mathcal{L}^b_\mathcal{Z}(\theta_g; z, \theta_e) =  
    \frac{1}{2} \left\Vert\frac{\mu_{\theta_e}(x) - \sqrt{1-\sigma^2_{\theta_e}}z }{\sigma_{\theta_e}}\right\Vert^2
\end{equation}
With this loss the optimal solution $\hat{x}^*(z)$ verifies $\mu_{\theta_e}(\hat{x}^*(z)) = \sqrt{1-\sigma^2_{\theta_e}}z$ thus $p(\mu_{\theta_e}(\hat{x}^*(z))) = \mathcal{N}(\mu_{\theta_e}(x);0, I-\Sigma) = p(\mu_{\theta_e}(x))$ as we have seen, its ensure that this loss has a common optimum with the GAN objective. This new loss takes into account the fact that when $\sigma_{\theta_e}$ is large, the observed $z$ is mostly noise and $\mu_{\theta_e}(x)$ is close to zero. The loss resulting from these considerations is $\mathcal{L}_G = \mathcal{L}^b_{\mathcal{Z}} + \mathcal{L}_{\mathcal{M}}$. It combines a GAN type loss $\mathcal{L}_\mathcal{M}$ and a reconstruction loss on the latent codes $\mathcal{L}_\mathcal{Z}$ which is similar to that described in Section~\ref{sec:complementarity}. The AVAE framework is globally presented in Figure~\ref{fig:framework}, with the relations between its components, and Figure~\ref{alg:method} gives the algorithm to train it. From a GAN perspective, the method can be viewed as constraining the latent space organization of the generator with the encoder model. It thus limits to some point the problem of mode collapse as the reconstruction error on the latent code prevents the generator to produce similar samples. As a consequence, it counteracts the mechanism pointed out by \cite{UNROLLEDGAN} to explain mode collapse by pushing generated samples apart from each other. The proposed architecture differs from VAE/GAN on several important aspects. The decoder and generator are separated in our work and our reconstruction error is based on the encoder model and not on the discriminator as in VAE/GAN to ensure that the error is informative about the image content. 
\begin{figure}[t]
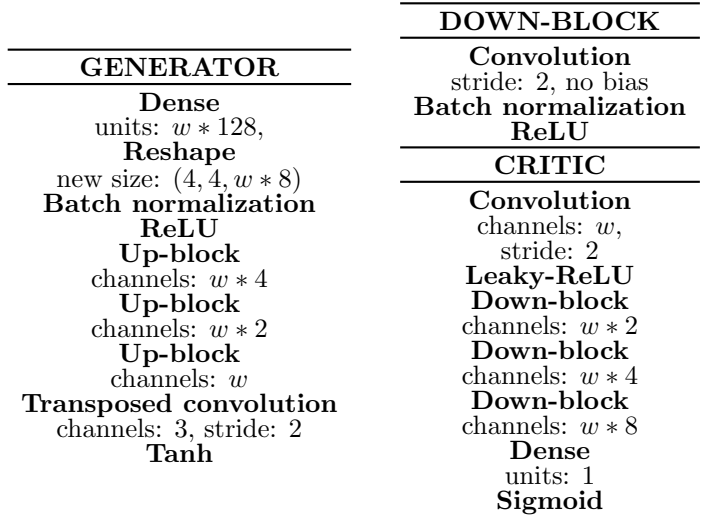

\centering
\setlength\tabcolsep{5pt}
\def\arraystretch{0.2}
\begin{tabular}{cc}
    \def\arraystretch{0.7}
    \begin{tabular}{c}
        \toprule
        \textbf{GENERATOR} \\
        \toprule
        \textbf{Dense} \\
        units: $w * 128$, \\  
        % \midrule
        \textbf{Reshape} \\
        new size: $(4, 4, w * 8)$ \\
        % \midrule
        \textbf{Batch normalization} \\
        % \midrule
        \textbf{ReLU} \\
        % \midrule
        \textbf{Up-block} \\
        channels: $w * 4$ \\
        % \midrule
        \textbf{Up-block} \\
        channels: $w * 2$ \\
        % \midrule
        \textbf{Up-block} \\
        channels: $w$  \\
        % \midrule
        \textbf{Transposed convolution} \\
        channels: $3$, stride: $2$ \\
        % \midrule
        \textbf{Tanh} \\
        % \bottomrule
    \end{tabular}
    &
    \begin{tabular}{c}
        \def\arraystretch{0.7}
        \begin{tabular}{c}
            \toprule
            \textbf{DOWN-BLOCK} \\
            \toprule
            \textbf{Convolution} \\
            stride: $2$, no bias \\
            % \midrule
            \textbf{Batch normalization} \\
            % \midrule
            \textbf{ReLU} \\
            \toprule
            \textbf{CRITIC} \\
            \toprule
            \textbf{Convolution} \\
            channels: $w$, \\
            stride: $2$ \\  
            % \midrule
            \textbf{Leaky-ReLU} \\
            % \midrule
            \textbf{Down-block} \\
            channels: $w * 2$ \\
            % \midrule
            \textbf{Down-block} \\
            channels: $w * 4$ \\
            % \midrule
            \textbf{Down-block} \\
            channels: $w * 8$ \\
            % \midrule
            \textbf{Dense} \\
            units: 1 \\
            % \midrule
            \textbf{Sigmoid} \\
            % \bottomrule
        \end{tabular}
        % \\
        % \def\arraystretch{0.7}
        % \begin{tabular}{c}
        %     \toprule
        %     \textbf{DOWN-BLOCK} \\
        %     \toprule
        %     \textbf{Convolution} \\
        %     stride: $2$, no bias \\
        %     % \midrule
        %     \textbf{Batch normalization} \\
        %     % \midrule
        %     \textbf{ReLU} \\
        %     % \bottomrule \\
        %     % ~ \\
        % \end{tabular}
    \end{tabular}
\end{tabular}
\caption{Generator and critic architectures. The decoder architecture is identical to the generator architecture and the encoder architecture differs from the critic architecture by the number of units in the last layer and by the absence of a Sigmoid activation at the end. Up-block is similar to Down-block but with transposed convolutions instead of convolutions and ReLUs instead of leaky-ReLUs. All convolutions and transposed convolutions share the same filter size ($5\time 5$) and use `same` padding. $\sigma_z$ is chosen independent of $x$ and is learned directly. $w$ is a width multiplier (we typically use $w = 128$). For the BiGAN implementation, we use a two-hidden-layer MLP for the latent code inputs and a critic-style architecture for the image inputs. The two outputs representations are then concatenated and used as input of a two-hidden-layers MLP.}
\label{tab:architecture}
\vskip -0.2in
\end{figure}
\section{Experimental results}
\textbf{Datasets}: We evaluate the models on six image datasets: LSUN bedroom \cite{LSUN} (64x64 images of bedrooms), CelebA \cite{CELEBA} (64x64 faces cropped images), FFHQ dataset (256x256 faces) \cite{STYLEGAN}, CIFAR10, CIFAR100 \cite{CIFAR} (32x32 images of 10 and 100 categories) and SVHN \cite{SVHN} (32x32 images of house numbers images). Images are resized to the sizes mentioned above and CelebA images are center-cropped at 70\%.

\textbf{Implementation details:} all the low resolution experiments have been conducted with Tensorflow 2.0 \cite{TENSORFLOW} on an NVIDIA GTX 1080 Ti GPU with 11Go of memory. Full code will be available on github. All models share similar architecture blocks, inspired by \cite{DCGAN}, to allow a fair comparison. Architecture details are presented in Figure~\ref{tab:architecture}. Each model is trained with hyper-parameters recommended in \cite{DCGAN} for $5e4$ iterations with a batch size of $64$. Because the reconstruction loss of the VAE part of VAE/GAN is a perceptual loss which differs from the MSE used in our model and in the classical VAE, the balance between the Kullback-Leibler divergence term and the reconstruction term in the VAE loss is not the same between models. We observed that the Kullback-Leibler divergence term is usually much higher for the VAE/GAN model which indicates that it conveys much more information in its latent code and thus introduces a bias in the reconstruction performance comparison between models. To solve this problem, we introduced a hyper-parameter $\beta$ to weight the Kullback-Leibler divergence in the encoder loss as in \cite{BETAVAE} in order to get similar Kullback-Leibler divergences. This hyper-parameter search leads us to the following ($\beta_\text{LSUN bedroom}=4$, $\beta_\text{celeba}=5$, $\beta_\text{CIFAR10}=10$, $\beta_\text{CIFAR100}=10$, $\beta_\text{SVHN}=20$). The high resolution experiment was conducted with a network architecture derived from the StyleGAN V2 architecture \cite{StyleGANV2} trained on 8 NVIDIA Quadro P5000 GPUs.
\begin{figure}[t]
\begin{center}
\vskip -0.1in
\includegraphics[width=1.0\columnwidth]{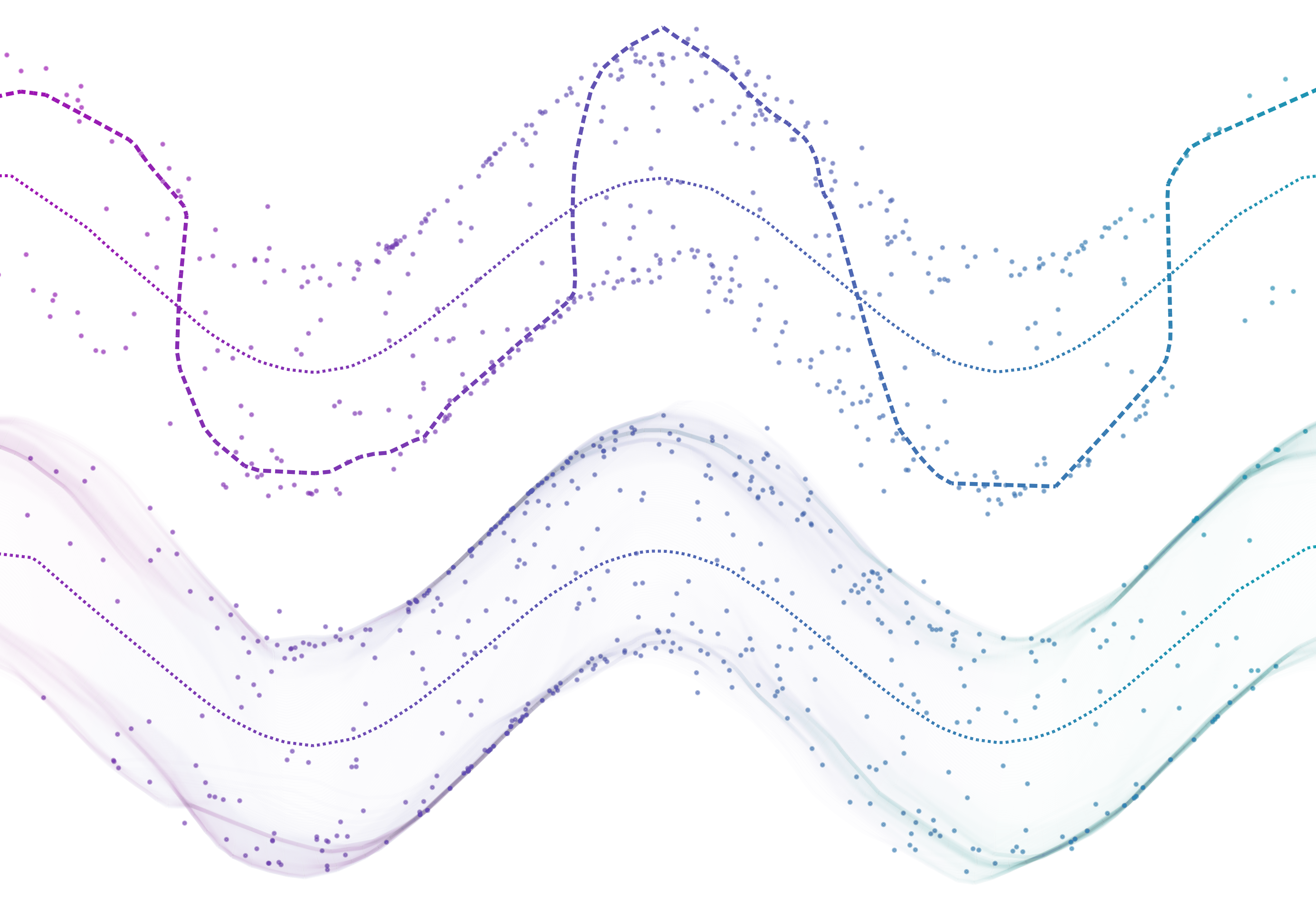}
\vspace{-0.3in}
\caption{Illustration of a toy example with two-dimensional data and a one-dimensional latent space. Points: data, dotted line: manifold of reconstructions from VAE, dashed line/density: manifold of reconstruction with our model. Color encodes the position in the one-dimensional latent space. Top: with a deterministic generator of the form $G_{\theta_g}(z)$. Bottom: with a probabilistic generator of the form $G_{\theta_g}(z, \xi)$. (best seen with zoom and color)}
\label{fig:manifold}
\end{center}
\vskip -0.3in
\end{figure}
\begin{figure}[t]
\begin{center}
\setlength\tabcolsep{2pt}
\begin{tabular}{Sr c}
real &
\cincludegraphics[width=0.78\columnwidth]{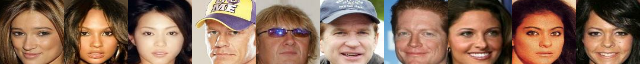}\\
VAE &
\cincludegraphics[width=0.78\columnwidth]{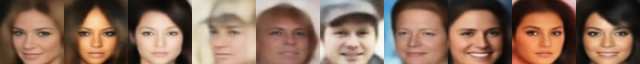}\\
VAE/GAN &
\cincludegraphics[width=0.78\columnwidth]{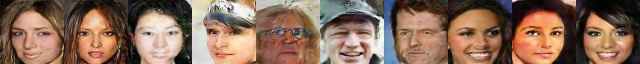}\\
BiGAN &
\cincludegraphics[width=0.78\columnwidth]{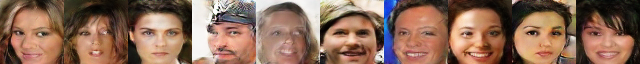}\\
Our &
\cincludegraphics[width=0.78\columnwidth]{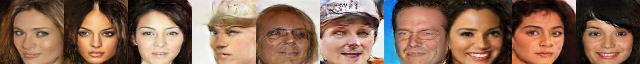}\\
\hline
real &
\cincludegraphics[width=0.78\columnwidth]{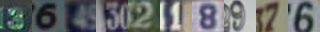}\\
VAE &
\cincludegraphics[width=0.78\columnwidth]{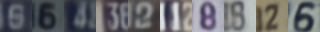}\\
VAE/GAN &
\cincludegraphics[width=0.78\columnwidth]{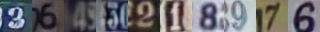}\\
BiGAN &
\cincludegraphics[width=0.78\columnwidth]{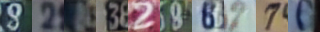}\\
Our &
\cincludegraphics[width=0.78\columnwidth]{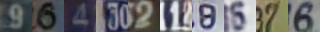}\\
\hline
real &
\cincludegraphics[width=0.78\columnwidth]{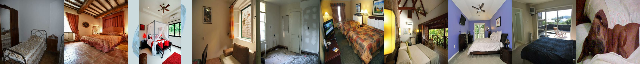}\\
VAE &
\cincludegraphics[width=0.78\columnwidth]{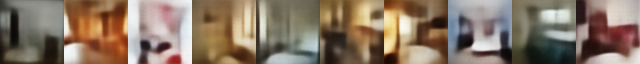}\\
VAE/GAN &
\cincludegraphics[width=0.78\columnwidth]{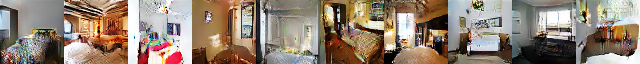}\\
BiGAN &
\cincludegraphics[width=0.78\columnwidth]{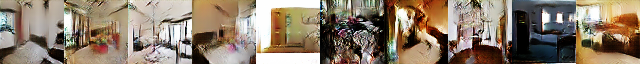}\\
Our &
\cincludegraphics[width=0.78\columnwidth]{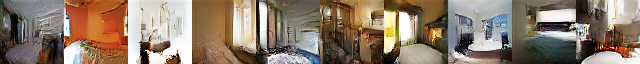}\\
\end{tabular}
\caption{Qualitative comparison of the quality of reconstructions between several frameworks namely VAE, VAE/GAN, BiGAN and our model on three datasets: CelebA, SVHN and LSUN bedroom.}
\label{fig:qualitative_rec}
\end{center}
\vskip -0.3in
\end{figure}
\begin{figure}
\vskip -0.3in
\begin{center}
\setlength\tabcolsep{2pt}
\begin{tabular}{Sr c}
real & 
\cincludegraphics[width=0.78\columnwidth]{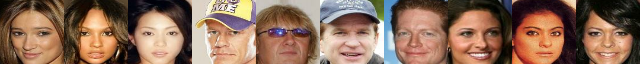}\\
VAE & 
\cincludegraphics[width=0.78\columnwidth]{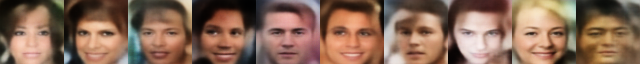}\\
GAN &
\cincludegraphics[width=0.78\columnwidth]{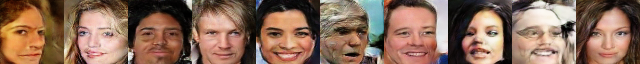}\\
VAE/GAN & 
\cincludegraphics[width=0.78\columnwidth]{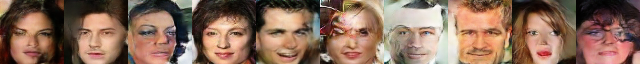}\\
BiGAN & 
\cincludegraphics[width=0.78\columnwidth]{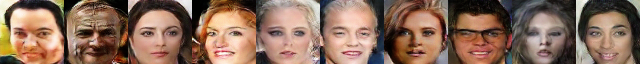}\\
Our &
\cincludegraphics[width=0.78\columnwidth]{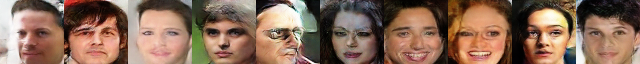}\\
\hline
real & 
\cincludegraphics[width=0.78\columnwidth]{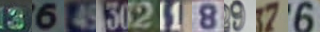}\\
VAE & 
\cincludegraphics[width=0.78\columnwidth]{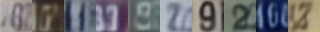}\\
GAN &
\cincludegraphics[width=0.78\columnwidth]{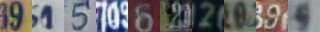}\\
VAE/GAN & 
\cincludegraphics[width=0.78\columnwidth]{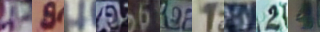}\\
BiGAN & 
\cincludegraphics[width=0.78\columnwidth]{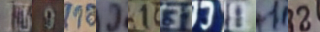}\\
Our &
\cincludegraphics[width=0.78\columnwidth]{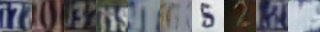}\\
\hline
real & 
\cincludegraphics[width=0.78\columnwidth]{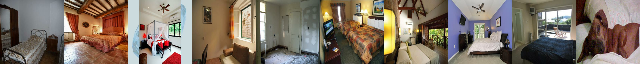}\\
VAE & 
\cincludegraphics[width=0.78\columnwidth]{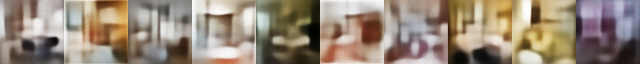}\\
GAN &
\cincludegraphics[width=0.78\columnwidth]{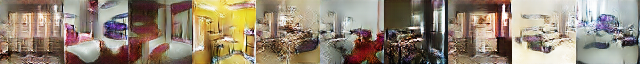}\\
VAE/GAN & 
\cincludegraphics[width=0.78\columnwidth]{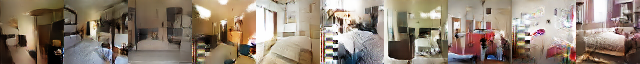}\\
BiGAN & 
\cincludegraphics[width=0.78\columnwidth]{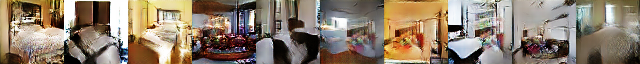}\\
Our &
\cincludegraphics[width=0.78\columnwidth]{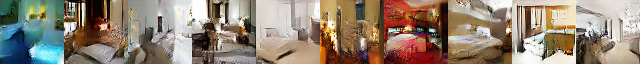}\\
\end{tabular}
\caption{Generated images for randomly sampled latent codes for CelebA, SVHN and LSUN bedroom.}
\label{fig:qualitative_gen}
\end{center}
\vskip -0.3in
\end{figure}
\begin{figure}[t]
\begin{center}
\setlength\tabcolsep{2pt}
\begin{tabular}{c c c}
\includegraphics[width=0.32\columnwidth]{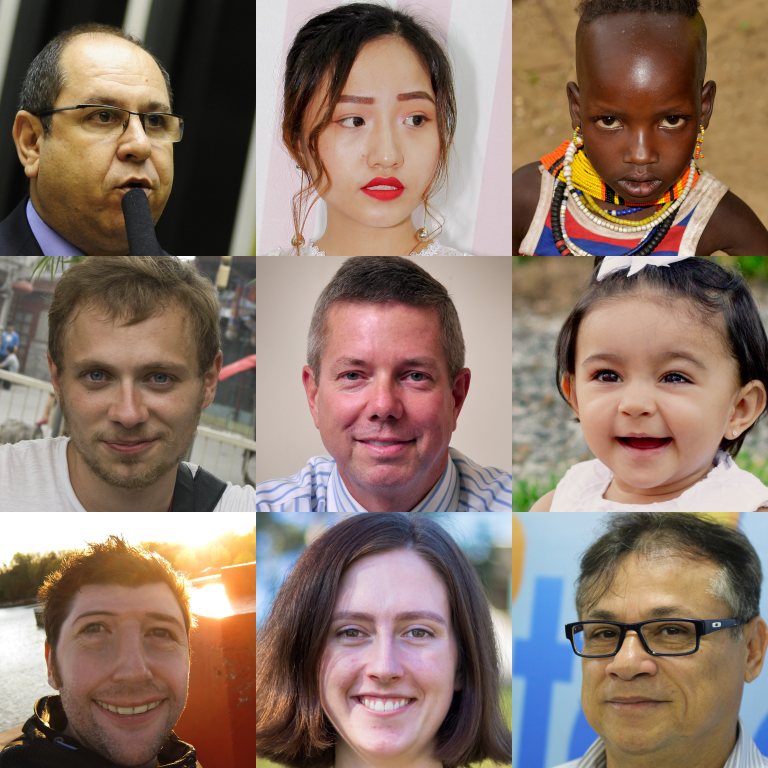} &
\includegraphics[width=0.32\columnwidth]{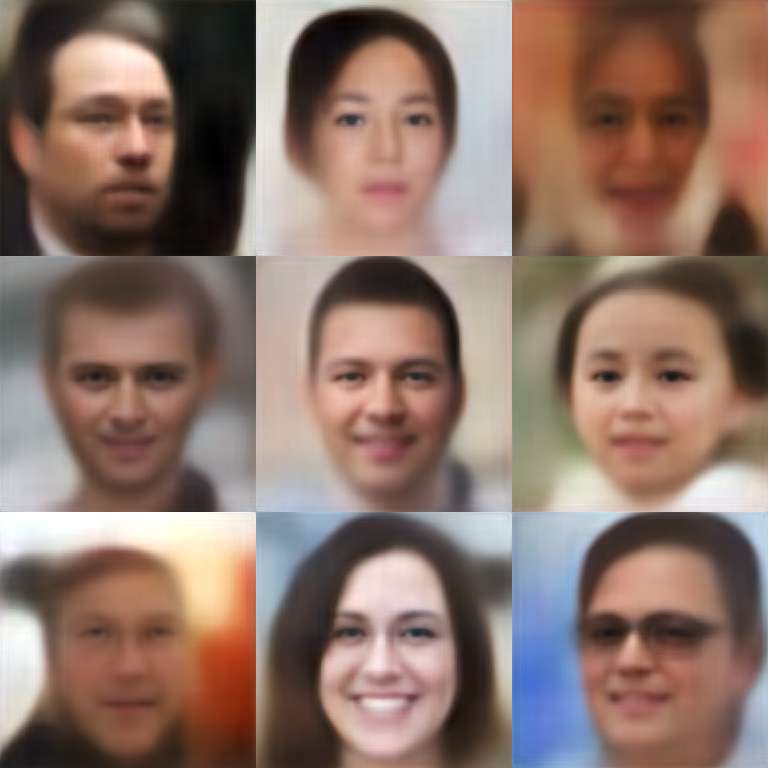} &
\includegraphics[width=0.32\columnwidth]{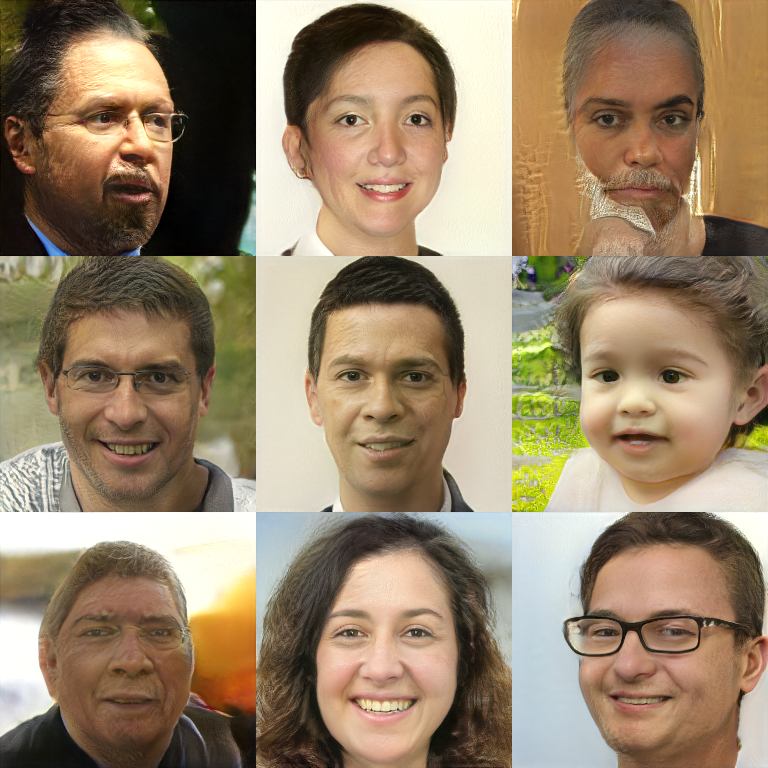} \\
\end{tabular}
\caption{Qualitative results on high resolution images. left to right: original images, reconstructions with the VAE decoder, reconstructions with the generator. This figure shows that with a limited amount of information the decoder fails to produce realistic reconstruction while our generator is capable of it. Note that the fidelity of the reconstruction is ultimately limited by the information contained in the latent code produced by the encoder.}
\label{fig:ffhq}
\end{center}
\vskip -0.3in
\end{figure}
\begin{table*}[t]
\caption{Reconstruction errors (MSE and LPIPS \cite{LPIPS}) and FID \cite{FID} of generated images for different models. Lower values are better for all metrics. Reported results are the average and standard deviation over five runs.}
\vskip -0.15in
\label{tab:quantitative results}
\begin{center}
\begin{small}
\begin{sc}
\setlength\tabcolsep{5pt}
\def\arraystretch{1.0}
\begin{tabular}{|cc|r|r|r|r|r|}
\cline{3-7}
\multicolumn{2}{c|}{} & Bedroom & CelebA & CIFAR10 & CIFAR100 & SVHN \\ 
\hline 
 & mse& $ 0.06\pm0.00 $& $ 0.03\pm0.00 $& $ 0.05\pm0.00 $& $ 0.05\pm0.00 $& $ 0.02\pm0.00 $ \\ 
VAE & lpips& $ 0.58\pm0.00 $& $ 0.18\pm0.00 $& $ 0.26\pm0.00 $& $ 0.25\pm0.00 $& $ 0.08\pm0.00 $ \\ 
 & fid& $229.75\pm1.45 $& $60.04\pm0.47 $& $136.75\pm0.57 $& $129.71\pm1.01 $& $68.16\pm2.10 $ \\ 
\hline 
GAN & fid& $110.59\pm19.55$& $14.54\pm0.41 $& $32.01\pm0.41 $& $34.51\pm0.59 $& $23.83\pm3.99 $ \\ 
\hline 
 & mse& $ 0.18\pm0.01 $& $ 0.07\pm0.00 $& $ 0.14\pm0.02 $& $ 0.15\pm0.02 $& $ 0.06\pm0.02 $ \\ 
VAE/GAN & lpips& $ 0.26\pm0.01 $& $ 0.09\pm0.00 $& $ 0.08\pm0.01 $& $ 0.08\pm0.01 $& $ 0.08\pm0.02 $ \\ 
 & fid& $60.02\pm2.36 $& $26.45\pm4.66 $& $39.04\pm2.42 $& $40.03\pm0.71 $& $17.02\pm2.58 $ \\ 
\hline 
 & mse& $ 0.42\pm0.05 $& $ 0.18\pm0.01 $& $ 0.31\pm0.02 $& $ 0.33\pm0.01 $& $ 0.12\pm0.01 $ \\ 
BiGAN & lpips& $ 0.44\pm0.02 $& $ 0.16\pm0.00 $& $ 0.14\pm0.00 $& $ 0.16\pm0.00 $& $ 0.12\pm0.01 $ \\ 
 & fid& $91.72\pm18.10$& $18.49\pm5.06 $& $34.61\pm1.29 $& $35.40\pm1.23 $& $27.77\pm2.96 $ \\ 
\hline 
Ours with $\xi$ & mse& $ 0.12\pm0.00 $& $ 0.05\pm0.00 $& $ 0.09\pm0.00 $& $ 0.09\pm0.00 $& $ 0.04\pm0.00 $ \\ 
with $\mathcal{L}^a_\mathcal{Z}$ & lpips& $ 0.36\pm0.00 $& $ 0.11\pm0.00 $& $ 0.10\pm0.00 $& $ 0.11\pm0.00 $& $ 0.10\pm0.00 $ \\ 
 & fid& $85.11\pm2.87 $& $16.99\pm0.58 $& $33.65\pm0.28 $& $39.81\pm0.60 $& $27.64\pm2.41 $ \\ 
\hline 
Ours without $\xi$ & mse& $ 0.12\pm0.00 $& $ 0.05\pm0.00 $& $ 0.09\pm0.00 $& $ 0.09\pm0.00 $& $ 0.04\pm0.00 $ \\ 
with $\mathcal{L}^a_\mathcal{Z}$  & lpips& $ 0.35\pm0.00 $& $ 0.11\pm0.00 $& $ 0.10\pm0.00 $& $ 0.11\pm0.00 $& $ 0.09\pm0.00 $ \\ 
 & fid& $84.29\pm5.28 $& $16.23\pm0.50 $& $33.49\pm0.50 $& $38.69\pm0.62 $& $28.47\pm8.24 $ \\ 
\hline 
Ours without $\xi$ & mse& $ 0.12\pm0.00 $& $ 0.05\pm0.00 $& $ 0.09\pm0.00 $& $ 0.09\pm0.00 $& $ 0.04\pm0.00 $ \\ 
with $\mathcal{L}^b_\mathcal{Z}$ & lpips& $ 0.35\pm0.00 $& $ 0.11\pm0.00 $& $ 0.10\pm0.00 $& $ 0.11\pm0.00 $& $ 0.08\pm0.00 $ \\ 
 & fid& $80.99\pm1.82 $& $15.01\pm0.82 $& $33.67\pm0.61 $& $38.35\pm0.57 $& $21.11\pm0.42 $ \\ 
\hline 
\end{tabular}
\end{sc}
\end{small}
\end{center}
\vskip -0.25in
\end{table*}
\subsection{Toy dataset}\label{sec:toy}
We begin by testing our approach on a toy dataset to validate the theory. The dataset is composed of 2D points generated from two generative factors $z_1$ and $z_2$. The data generation procedure is the following:
$z_1, z_2, \epsilon \sim \mathcal{N}(0, 1)$ and $x = f(z_1, z_2, \epsilon) = (3z_1 + 0.1\epsilon, cos(3 z_1) + \text{tanh}(3z_2) + 0.1\epsilon)$. For the model, we use a latent space of dimension one to simulate the problem of the low dimensionality of the latent space compared to the high dimensionality of the data manifold. Models are two-hidden-layer perceptrons with 128 units. Models are trained with the method described proposed in this paper. We then draw the manifold of the generated points to see how the model behave compared to a VAE. Results of this experiment can be seen in Figure~\ref{fig:manifold} where we can see that reconstructions from the VAE are in a region of low likelihood of the data distribution while AVAE reconstructions follow the shape of the VAE manifold while covering regions of higher likelihood. It shows that our model is able to produce realistic reconstructions even when the latent code do not contain all the information needed to reconstruct the original image perfectly. Here there is an ambiguity as we do not know if the original sample is from the top distribution or the bottom one given a latent code corresponds to two. In order to produce a realistic result the generator has to make an arbitrary choice. Our approach allows the generator to make such choice while the decoder from the VAE outputs the average of possible choices resulting in an unlikely/unrealistic reconstruction. On the same Figure we can see that when using a stochastic generator with additional latent variables, it learns to generate missing regions of the data distribution while keeping the VAE latent space structure. 
\subsection{Qualitative results}
Here, we present some qualitative results on the CelebA SVHN and LSUN bedroom datasets. A comparison of samples reconstruction between our model and other models is presented in Figure~\ref{fig:qualitative_rec}. We also present a visual comparison of samples generated by our model and other generative models in Figure~\ref{fig:qualitative_gen}. Additional qualitative results will be available on github. We can see on these figures that generated images are of comparable quality of GAN generated images for both generation and reconstructions. VAE reconstructions and generated samples look blurry, BiGAN generated images are of good quality but reconstructions are not accurate. VAE/GAN produces both good reconstructions and generated samples. However, while our judgment is subjective, we find that reconstructions produced by VAE/GAN are less accurate than ours and images are less realistic than with GAN, BiGAN or our approach. 

One may notice that for the LSUN bedroom dataset, reconstructions produced by our model are not convincing. However, we can explain this by the very poor performance of the VAE suggesting that not enough information passes through the latent code to create a reconstruction visually close to the original image. However, even here, our model still produces sharp images close to the target ones in terms of MSE showing that our model follows the latent structure of the VAE trained with the MSE as a reconstruction error. 

We also conducted an experiment on higher resolution images (256x256 FFHQ face images) to see if our method can be scaled to high resolution images. To conduct this experiment, we made straightforward modifications to the style GAN V2 \cite{StyleGANV2} using the approach proposed here. Results of this experiments are presented in Figure~\ref{fig:ffhq}. These results confirm the scalability of the proposed approach to bigger architectures. 
\subsection{Quantitative results}
To quantitatively evaluate the performance of our method, we selected several metrics. The quality of the reconstructed images is evaluated by the Mean Squared Error or MSE and the LPIPS \cite{LPIPS}. We use the FID \cite{FID} to measure the realism of generated images. A comparison between VAE, GAN, BiGAN \cite{ALI, BIGAN}, VAE/GAN \cite{VAEGAN}, and our model is presented in Table~\ref{tab:quantitative results}. Reconstructions errors are computed on validation images not used during training, namely the \emph{test} or \emph{validation} splits of TensorFlow datasets. FID is computed over 50000 randomly generated samples and compared to training data samples as FID requires a lot of samples to be calibrated. It must be noted that some metrics are biased toward some architectures: the MSE is favorable to the VAE model because it is the loss used to train it. It is also the case for our approach, as information contained in the latent code is optimized to produce accurate reconstructions in terms of MSE. VAE/GAN is also advantaged in terms of LPIPS and FID as this model uses a perceptual similarity metric based on a classifier as a reconstruction error and the FID and LPIPS are also based on deep features. Globally, our model exhibits a good compromise between accurate reconstructions (MSE and LPIPS) and realism (FID), thus combining the best of VAE and GAN.
\section{Discussion}
The proposed framework can be used to generate images from a pre-trained representation. Thus, it is not a feature learning method and only features learned by the VAE are described by the representation. However, while we focused on a VAE architecture to produce the latent representation, our approach can be further extended. Indeed one could for example train a classifier while constraining its last feature layer in the same way the latent code is constrained and use it as a latent code in our method in order to focus on different features of the image. One could even concatenate several of these representation to train a model which fits their needs. We keep this extension as a potential future work.

\textbf{Acknowledgment} This paper was supported by European Union´s Horizon 2020 research and innovation programme under grant number 951911 - AI4Media. This publication was made possible by the use of the FactoryIA supercomputer, financially supported by the Ile-de-France Regional Council

\bibliography{references}
\bibliographystyle{plain}
\end{document}